%% file: main.tex
\documentclass[10pt,twocolumn,letterpaper]{article}

\usepackage[pagenumbers]{cvpr} 

\input{preamble}

\definecolor{cvprblue}{rgb}{0.21,0.49,0.74}
\usepackage[pagebackref,breaklinks,colorlinks,citecolor=cvprblue]{hyperref}
\usepackage{multirow}
\usepackage{adjustbox}

\usepackage{xspace}
\makeatletter 
\DeclareRobustCommand\onedot{\futurelet\@let@token\@onedot} 
\def\@onedot{\ifx\@let@token.\else.\null\fi\xspace}  
\def\eg{\emph{e.g}\onedot} 
 
\def\ie{\emph{i.e}\onedot}

\makeatother 

\usepackage[T1]{fontenc}

\def\paperID{3288} 
\def\confName{CVPR}
\def\confYear{2025}

\title{VIRES: Video Instance Repainting via Sketch and Text Guided Generation}

\author{Shuchen Weng$^{1,2^{\dag\ddag}}$ \quad
Haojie Zheng$^{3,4^{\dag}}$ \quad
Peixuan Zhang$^{5}$ \quad
Yuchen Hong$^{1,2}$ \quad \\
Han Jiang$^{3}$ \quad
Si Li$^{5}$ \quad 
Boxin Shi$^{1,2^{*}}$ \quad
\vspace{0.3em} \\
\normalsize \textsuperscript{1} National Key Laboratory for Multimedia Information Processing, School of Computer Science, Peking University \\
\normalsize \textsuperscript{2}  National Engineering Research Center of Visual Technology, School of Computer Science, Peking University \\
\normalsize   \textsuperscript{3} OpenBayes Information Technology Co., Ltd.  \qquad \textsuperscript{4} School of Software and Microelectronics, Peking University\\
\normalsize  \textsuperscript{5} School of Artificial Intelligence, Beijing University of Posts and Telecommunications
}

\begin{document}

\twocolumn[{%
\renewcommand\twocolumn[1][]{#1}%
\maketitle

\begin{center}
    \vspace{-20pt} 
    \centering
    \captionsetup{type=figure}
    \includegraphics[width=\linewidth]{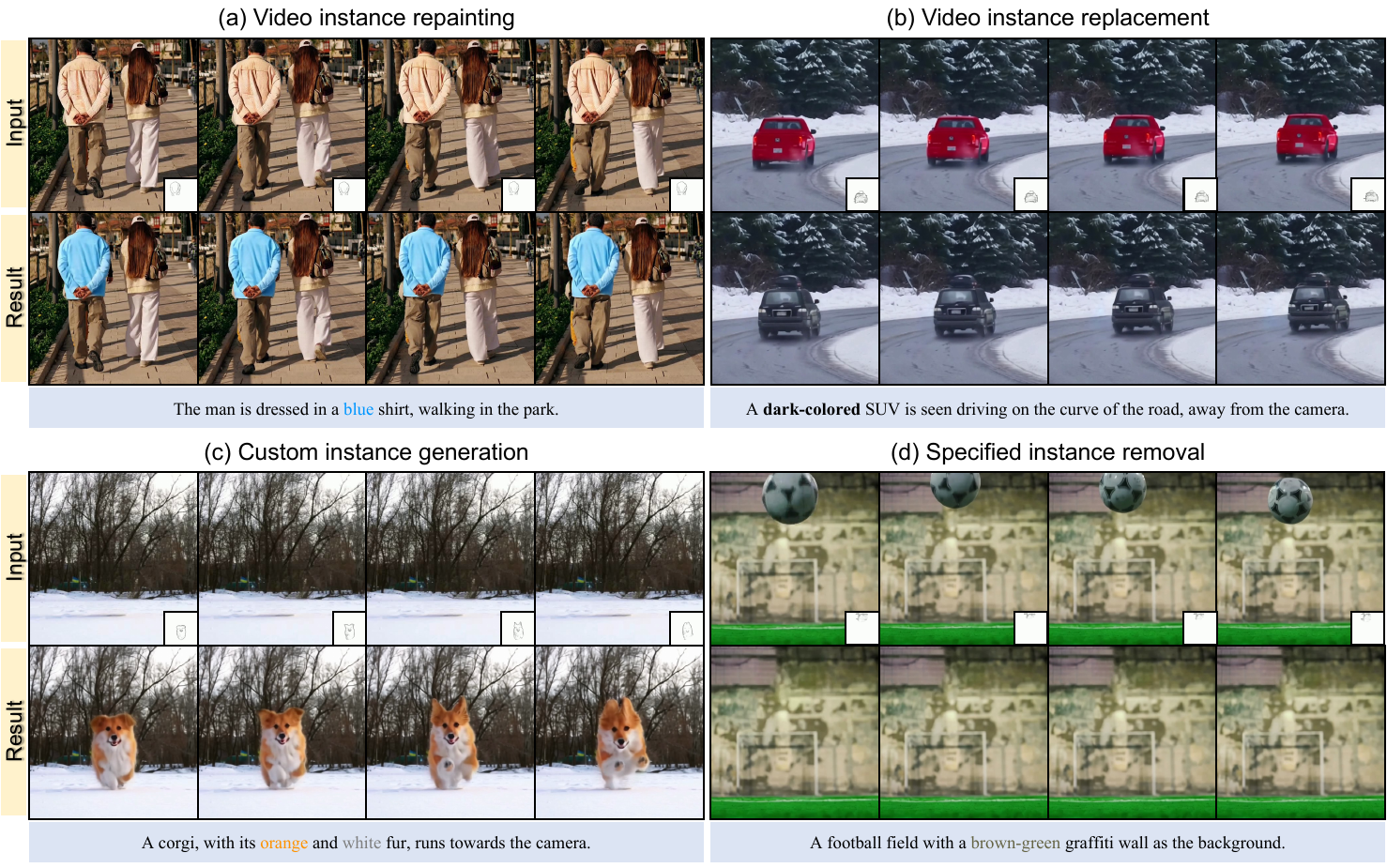}
    \vspace{-20pt} 
   \captionof{figure}{Our VIRES model demonstrates powerful video editing capabilities with sketch and text guidance, as shown in four typical scenarios: (a) Repainting the color and style of the man's shirt. (b) Replacing the pickup truck with the dark-colored SUV.  (c) Generating a running corgi within a video clip. (d) Removing a specified football from a video clip. }
    \label{fig:teaser}
\end{center}
}]

\let\thefootnote\relax
\footnotetext{
    \begin{minipage}[t]{\textwidth}
        $^\dag$ Equal contribution. \\
        $^\ddag$ Now at Beijing Academy of Artificial Intelligence. \\
        $^*$ Corresponding author.
    \end{minipage}
}

\begin{abstract}
We introduce VIRES, a video instance repainting method with sketch and text guidance, enabling video instance repainting, replacement, generation, and removal. Existing approaches struggle with temporal consistency and accurate alignment with the provided sketch sequence. VIRES leverages the generative priors of text-to-video models to maintain temporal consistency and produce visually pleasing results. We propose the Sequential ControlNet with the standardized self-scaling, which effectively extracts structure layouts and adaptively captures high-contrast sketch details. We further augment the diffusion transformer backbone with the sketch attention to interpret and inject fine-grained sketch semantics. A sketch-aware encoder ensures that repainted results are aligned with the provided sketch sequence. Additionally, we contribute the {\sc VireSet}, a dataset with detailed annotations tailored for training and evaluating video instance editing methods. Experimental results demonstrate the effectiveness of VIRES, which outperforms state-of-the-art methods in visual quality, temporal consistency, condition alignment, and human ratings. Project page:\href{https://hjzheng.net/projects/VIRES/}{https://hjzheng.net/projects/VIRES/}

\vspace{-10mm}

\end{abstract}

\vspace{4mm}
\section{Introduction}
Video is a crucial medium for people to capture and communicate their experiences and ideas. 
Traditionally, maintaining temporal consistency when editing videos requires specialized skills and considerable effort.
With the advancements in diffusion models~\cite{ddim, ddpm}, recent text-guided video editing methods~\cite{Tune-a-video, Fatezero} attract considerable interest and offer the potential to revolutionize filmmaking and entertainment through a controllable approach.

While text can convey general video editing goals, it often leaves considerable room for interpretation of the user's intent (\eg, logo details on clothing). As the improvement, recent works~\cite{vidtome, renderavideo, text2videozero} introduce additional guidance (\eg, sketch) for more precise control. However, relying on pre-trained text-to-image models (\eg, Stable Diffusion \cite{stablediffusion}), these zero-shot methods struggle with temporal consistency, inevitably resulting in flickering.
As an alternative, VideoComposer~\cite{videocomposer} fine-tunes the text-to-image model and incorporates temporal modeling layers. However, its focus on compositionality limits the accurate alignment with the provided sketch sequence, resulting in suboptimal fine-grained instance-level edits.




In this paper, we propose the \textbf{VIRES}, a \textbf{V}ideo \textbf{I}nstance \textbf{RE}painting method with \textbf{S}ketch and text guidance. As illustrated in~\cref{fig:teaser}, VIRES facilitates four application scenarios to repaint instances within the provided video (\ie, video instance repainting, replacement, generation, and removal).
Leveraging the generative priors of text-to-video models (\ie, Open-Sora~\cite{opensora}), our approach maintains temporal consistency and produces visually pleasing results.
To provide precise user control over instance properties, we design tailored modules that process sketch for the structure (\eg, pose and movement) and text for the appearance (\eg, color and style). 
Specifically, we present the Sequential ControlNet to effectively extract structure layouts and the standardized self-scaling to adaptively capture high-contrast structure details. 
Additionally, to interpret and inject fine-grained sketch semantics into the latent space, we augment the Diffusion Transformer (DiT) backbone with the sketch attention. 
Finally, the sketch-aware encoder aligns repainted results with multi-level texture features during the latent code decoding.


We further contribute the {\sc VireSet}, a new dataset for training and evaluating video instance editing methods. {\sc VireSet} provides 85K training videos and 1K evaluation videos, sourced from the SA-V~\cite{sam2}. For each video, we generate text annotations from the large language model (LLM) \cite{pllava} and extract sketch sequences using HED edge detection~\cite{hed}. Our contributions are summaries as follows:
\begin{itemize}
    \item We propose the first DiT-based framework for controllable video instance editing, along with a tailored dataset for training and evaluating relevant editing methods.
    \item We present the Sequential ControlNet and the standardized self-scaling to effectively extract structure layouts and adaptively capture high-contrast sketch details.
    \item We introduce the sketch attention to interpret and inject fine-grained sketch semantics and sketch-aware encoder to align repainted results with multi-level texture features.
\end{itemize}

\section{Related work}

\subsection{Diffusion-based image generation}
Diffusion models~\cite{ddim, ddpm} have demonstrated remarkable advancements in image generation~\cite{stablediffusion, imagen}. These models learn to reverse the forward diffusion process, generating images by iterative denoising a sample from a standard Gaussian distribution. To control this generation process, researchers use adapter mechanisms to incorporate additional conditions~\cite{controlnet, t2iadapter}. 
Leveraging the generative priors encapsulated in the pre-trained generation model, users can customize content via fine-tuning~\cite{dreambooth, multiconcept} and edit arbitrary images via training-free strategies~\cite{sdedit, pix2pixzero}. 
These advancements expand the application of diffusion models to various image-based tasks, \eg, super resolution~\cite{SinSR}, image colorization~\cite{l-cad}, and reflection removal~\cite{L-DiffER}. 
Given the similar spatial representations between images and videos, researchers are inspired to effectively lift these pre-trained image priors to the video domain.

\subsection{Diffusion-based video generation}
Early text-to-video generation works attempt to extend the UNet-based denoising network with space-only 3D convolutions, training them from scratch~\cite{vdm}. 
However, this approach proves computationally expensive, leading researchers~\cite{genl, lvdm} to explore fine-tuning pre-trained text-to-image models~\cite{stablediffusion}. This fine-tuning typically introduces additional temporal modules (\eg, temporal convolutions~\cite{i3d} or temporal attention~\cite{timesformer}) to improve the temporal consistency, while freezing most of the pre-trained denoising network weights to preserve spatial generative priors.
To further reduce generation costs, researchers explore zero-shot text-guided video editing that solely performs during the inference phase~\cite{text2videozero, ControlVideo}.
Recently, impressive text-to-video generation results from OpenAI~\cite{sora} have renewed interest in training models from scratch, but this time leveraging the diffusion transformer architecture~\cite{opensora, allegro, cogvideox}.
Given the significant improvements demonstrated on video generation benchmarks~\cite{vbench}, we build upon this foundation and develop modules for controllable video generation.


\section{Dataset} \label{sec:dataset}
Existing datasets~\cite{davis, WebVid10M} face challenges in providing precise instance masks due to temporal inconsistencies (\eg, occlusions and reappearances) and variations in visual appearance (\eg, complex motion, lighting, and scale).  Therefore, we present the {\sc VireSet}, a dataset including high-quality instance masks tailored for training and evaluating video instance editing methods. 

\noindent \textbf{Data acquisition.} Our initial video samples are sourced from SA-V~\cite{sam2}, a dataset includes indoor (54\%) and outdoor (46\%) scenes recorded across 47 countries by diverse participants. The original videos have an average duration of 14 seconds and a resolution of $1401 \times 1037$ pixels, with multiple manual and automatic instance mask annotations provided at 6 FPS. 
SA-V adopts a data engine with a verification step to ensure high-quality instance mask annotations. To further improve temporal consistency for video instance editing, we leverage the pre-trained SAM-2 model~\cite{sam2} to annotate the intermediate frames, increasing the mask annotation rate to 24 FPS.

\noindent \textbf{Instance selection.} We filter instances to include only those covering at least 10\% of the frame area. Additionally, We only use instances present for at least 51 consecutive frames, from which we randomly sample 51-frame clips. Each selected instance is then cropped from the original video with a small margin around its bounding box and resized to $512 \times 512$ resolution.

\noindent \textbf{Additional annotations.} We adopt the pre-trained PLLaVa~\cite{pllava} to generate text descriptions for each cropped video clip, capturing their visual appearance. To ensure the quality, we recruit 10 volunteers to review 1\% samples, achieving a 91\% acceptance rate.
Following VideoComposer~\cite{videocomposer}, we extract sketch sequences using HED edge detection~\cite{hed} to provide the structure guidance.

\noindent \textbf{Dataset statistics.} The final {\sc VireSet} includes 85K video clips for training and 1K video clips for evaluation. Each clip consists of 51 frames at 24 FPS with a resolution of $512 \times 512$, accompanied by a sketch sequence for the structure and a text for the appearance.



\begin{figure*}[t]
   \centering
  \includegraphics[width=\linewidth]{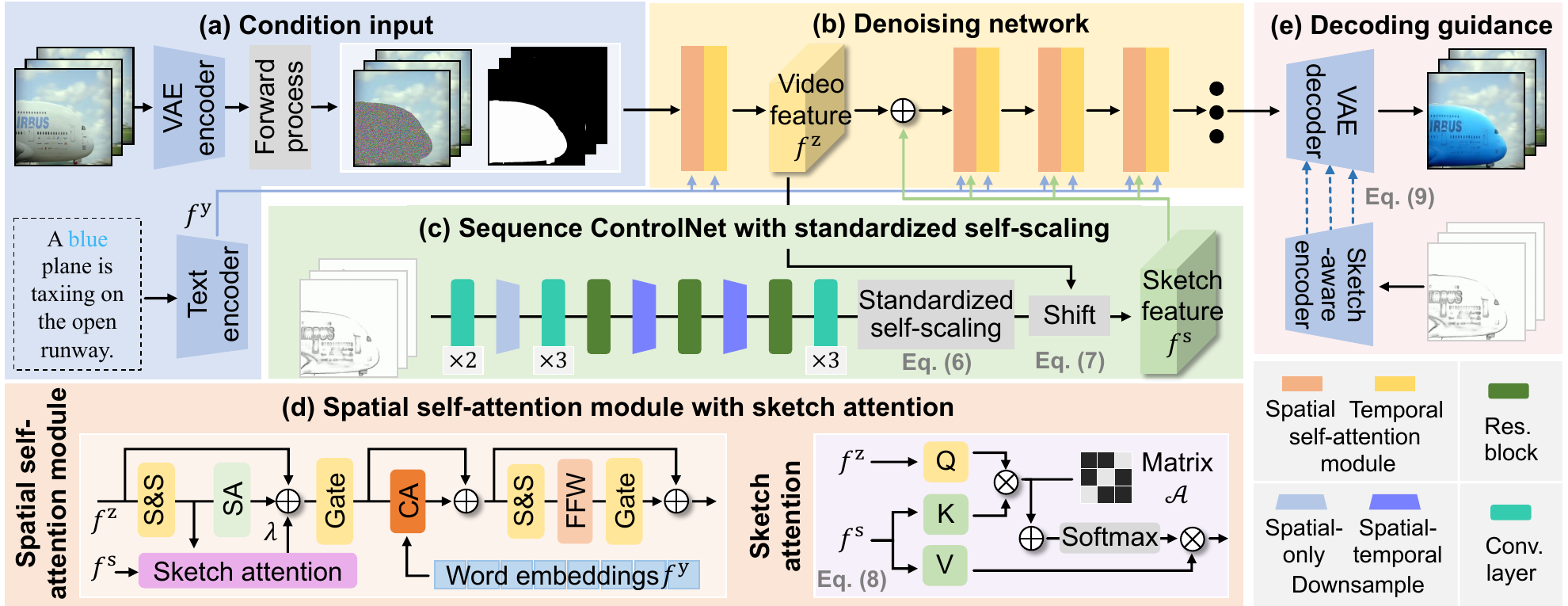}
  \caption{The pipeline of VIRES model.
  (a) The input video clip and corresponding text descriptions are encoded using independent encoders. 
  Noise is selectively added to the latent code according to the instance mask.  
  (b) This noised latent code is fed into the denoising network, composed of staked transformer blocks with spatial and temporal self-attention modules, trained to reverse the diffusion process.  
  (c) The Sequential ControlNet extracts structure layouts and injects them into the denoising network with the standardized self-scaling. 
  (d) The sketch attention is implemented as a parallel branch within the spatial self-attention module, injecting structure semantics into subsequent transformer blocks.
  (e) A sketch-aware encoder additionally provides multi-level texture features during decoding, generating the repainted video clip from the denoised latent code.
  }
  \label{fig:pipeline}
\end{figure*}

\section{Methodology}
This section begins with an overview of the framework (\cref{sec:overview}). We then delve into the proposed modules: the Sequential ControlNet with the standardized self-scaling (\cref{sec:extraction}), the sketch-based DiT backbone with the sketch attention (\cref{sec:interpretation}), and the sketch-aware encoder to improve the decoding process (\cref{sec:vae}).

\subsection{Overview} \label{sec:overview}
Given an $N$-frame original video clip $x = \{ x^i \}_{i=1}^N$, a corresponding sketch sequence $s = \{ s^i \}_{i=1}^N$ presenting the structure, a text description $y$ describing the appearance, and an instance mask sequence $m = \{ m^i \}_{i=1}^N$ indicating the repainted regions, VIRES is proposed to repaint the specific video instance. The overall pipeline is illustrated in~\cref{fig:pipeline}.

\noindent \textbf{Input encoding.} \label{sec:input}
The original video clip $x$ is encoded into the latent code $z$ using the pre-trained spatial-temporal VAE encoder $\mathcal{E}^{\mathrm{x}}$~\cite{opensora} as $z = \mathcal{E}^{\mathrm{x}}(x)$. 
The sketch sequence $s$ is processed by our proposed Sequential ControlNet $\mathcal{F}^{\mathrm{s}}$ to effectively extract structure layouts as $f^{\mathrm{s}} = \mathcal{F}^{\mathrm{s}}(s)$.
The text $y$ is encoded with the pre-trained text encoder $\mathcal{F}^{\mathrm{y}}$~\cite{t5} to obtain the word embeddings as $f^{\mathrm{y}} = \mathcal{F}^{\mathrm{y}}(y)$.

\noindent \textbf{Condition injection.} \label{sec:injection}
To adaptively capture the high-contrast structure details of the sketch sequence, we introduce the standardized self-scaling.
We further augment the DiT backbone with the sketch attention, which interprets and injects fine-grained sketch semantics into the latent space.
Word embeddings are injected into the DiT backbone with the pre-trained cross-attention modules. 


\noindent \textbf{Forward process.} We adopt the Flow Matching formulation~\cite{flowmatching} for robust and stable training. 
For simplicity, we consider a linear path between the sampled latent code $z$ and the Gaussian noise $\epsilon \sim \mathcal{N}(0, 1)$:
\begin{equation}
    \hat{z}_t = tz + (1 - t)\epsilon,
\label{eq:add_noise}
\end{equation}
where $t \in [0, 1]$ is the timestep, with $\hat{z}_0 = \epsilon$ and $\hat{z}_1 = z$.

\noindent \textbf{Latent masking.}
After concatenating the mask with the latent code, we further selectively add noise according to the instance mask, allowing the model to repaint specific instances within the video:
\begin{equation}
z_t = \hat{z}_t \odot \hat{m} + z \odot (1 - \hat{m}),
\end{equation}
where $\hat{m}$ is the downsampled mask sequence indicating where the repainted instance places. 

\noindent \textbf{Backward process.} \label{sec:backward}
The reverse diffusion process can be defined by the Ordinary Differential Equation (ODE):
\begin{equation}
    dz/dt = v_\theta(z_t, t, s, m, y),
\end{equation}
where $v_\theta(z_t, t, s, m, y)$ is the estimated vector field that guides the generation process.
By solving the ODE from $t=0$ to $t=1$, we transform the noise into the latent code.

\noindent \textbf{Denoising learning.}
We adopt the flow matching objective~\cite{flowmatching} to train the denoising network $v_\theta (z_t,t, s, m, y)$. In the case of the linear path, the target velocity $v_t$ simplifies to $z - \epsilon$. We optimize the estimated velocity $v_\theta(z_t, t, s, m, y)$ towards the target velocity $v_t$ by minimizing the loss:
\begin{equation}
\mathcal{L}_\mathrm{fm} =  \mathbb{E}_{t,z,\epsilon} \big[ \left\| v_t(z_t) - v_\theta(z_t, t, s, m, y) \right\|^2 \big].
\end{equation}



\noindent \textbf{Latent decoding.}
The denoised latent code $z^\prime$ is then decoded into a video clip $\bar{x}$ using the pre-trained spatial-temporal VAE decoder $\mathcal{D}$~\cite{opensora}. To further align the repainted results with the sketch, we introduce a sketch-aware encoder $\mathcal{E}^{\mathrm{s}}$, which adopts the same VAE encoder architecture to provide multi-level texture features as the guidance during decoding:
\begin{equation}
    \bar{x} = \mathcal{D}(z^\prime,\mathcal{E}^{\mathrm{s}}(s)).
\end{equation}

\subsection{Sequential sketch feature extraction} \label{sec:extraction}
As presented in~\cref{fig:pipeline} (c), we propose the Sequential ControlNet to extract structure layouts from the sketch sequence. This is followed by the standardized self-scaling, designed to adaptively capture the high-contrast structure details of the extracted features.

\noindent \textbf{Sequential ControlNet.}
Previous ControlNet~\cite{controlnet} is designed for image editing, which lacks temporal consistency and can produce flickering artifacts when applied to video-based tasks. This motivates us to introduce Sequential ControlNet\footnote{Further variations are discussed in the Supp.} for the video instance repainting. 
Our Sequential ControlNet includes convolutional layers, residual blocks, and downsampling layers. Each convolutional layer consists of a 3D causal convolution~\cite{3dcausal}, followed by a Group Normalization~\cite{groupnorm} and a SiLU activation function~\cite{silu}, to effectively capture spatial-temporal dependencies between frames. 
Residual blocks consist of two such convolutional layers with a residual connection. Downsampling layers are convolutional layers with a stride of 2, applied to compress the spatial or spatial-temporal dimensions. 
Since the spatial-temporal VAE encoder~\cite{opensora} downsamples the original video clip by $4 \times$ temporally and $64 \times$ spatially, we adopt one spatial downsampling layer and two spatial-temporal downsampling layers to match the feature map dimensions of the DiT backbone.
To align the embedding channels, we progressively increase the number of channels. Specifically, we increase the embedding channels in the shallow convolutional layers and double the channels at each downsampling layer. 
The final three layers maintain a high channel dimension to effectively extract structure layouts from the sketch sequence.

\noindent \textbf{Standardized self-scaling.}
Feature modulation has proven effective in conditional image editing (\eg, AdaIN~\cite{adain}, FiLM~\cite{film}, and SPADE~\cite{spade}). Observing that the sketch has high-contrast transitions between black lines and the white background, we introduce the standardized self-scaling to adaptively capture sketch details, instead of performing simply addition. Specifically, we use sketch features $f^\mathrm{s}$ extracted by the Sequential ControlNet and standardize them to scale the features themselves, effectively highlighting the high-contrast regions:
\begin{equation}
\begin{aligned}
\hat{f}^{\mathrm{s}} = \big((f^{\mathrm{s}} - \mu(f^{\mathrm{s}}))/\sigma(f^{\mathrm{s}})\big) \odot f^{\mathrm{s}}, 
\end{aligned}
\end{equation}
where $\mu(\cdot)$ and $\sigma (\cdot)$ represent the function of mean and standard deviation, respectively. We then shift the feature domain from sketch to video by aligning their means:
\begin{equation}
    \bar{f}^{\mathrm{s}} = \hat{f}^{\mathrm{s}} - \mu(\hat{f^{\mathrm{s}}}) + \mu(f^{\mathrm{z}}),
\end{equation}
where $f^{\mathrm{z}}$ represents the video features. To reduce computational cost, standardized self-scaling is applied only once to the first transformer block of the DiT backbone.



\subsection{Latent sketch semantic interpretation} \label{sec:interpretation}
As illustrated in~\cref{fig:pipeline} (d), following the standardized self-scaling that provides the high-contrast structure details, we further augment the DiT backbone with the sketch attention to interpret and inject fine-grained sketch semantics within the latent space. 

\noindent \textbf{DiT backbone.}
The VIRES framework builds upon the pre-trained text-to-video generation model~\cite{opensora}, leveraging its DiT architecture that consists of stacked transformer blocks. Each block incorporates separate spatial and temporal self-attention modules to capture intra-frame and inter-frame dependencies, respectively. In each module, Self-Attention (SA) and Feed-Forward Network (FFN) extract contextual features, Cross-Attention (CA) injects semantics from word embeddings, and the scale and shift (S\&S) and the gate modulate the latent code with timestep embeddings.
We concatenate the latent code with the instance mask before feeding it to the transformer, and patchify both the latent code and the sketch sequence into non-overlapping  $1 \times 2 \times 2$ tokens, resulting in a $4 \times$ spatial downsampling.

\noindent \textbf{Sketch attention.}
To interpret and inject sketch semantics into the latent space, we augment the DiT backbone with the sketch attention within each spatial self-attention module except for the first. The sketch attention incorporates a predefined binary matrix $\mathcal{A}$ to indicate correspondences between the latent code and the sketch sequence:
\begin{equation}
    \bar{f}^{\mathrm{z}} = \mathrm{Softmax}  \big( (QK^\top + \mathcal{A})/{\sqrt{C}} \big)V,
    \label{eq:attention}
\end{equation}
where $Q$ and $K, V$ are transformed features from the video features $f^{\mathrm{z}}$ and the extracted structure layouts $f^{\mathrm{s}}$, respectively. $C$ is the number of embedding channels.
Sketch attention is implemented as a parallel branch, and its outputs are added with a learnable scaling parameter $\lambda$, allowing adaptive weighting of injected sketch semantics.

\subsection{Visual sketch texture guidance} \label{sec:vae}
As illustrated in~\cref{fig:pipeline} (e), during decoding the latent code into a video clip, we additionally adopt a sketch-aware encoder, which provides multi-level texture features to the spatial-temporal VAE decoder and aligns the structure of repainted results with the provided sketch sequence.

\noindent \textbf{Spatial-temporal VAE.} 
We use the pre-trained spatial-temporal VAE~\cite{opensora} to encode the original video clip into a latent code and subsequently decode it back into a video clip. 
Specifically, the spatial-temporal encoder firstly uses a stack of 2D vanilla convolution blocks with three downsampling layers, achieving $64 \times$ spatially compression. Following this, 3D causal convolution blocks capture spatial-temporal dependencies with two downsampling layers applied only to the temporal dimension, resulting in $4 \times$ temporal compression.
The spatial-temporal VAE decoder mirrors the architecture of the encoder, using 3D causal convolutions for spatial-temporal modeling and upsampling the temporal dimension before the spatial dimension.


\noindent \textbf{Sketch-aware encoder.}
During decoding the latent code into a video clip, the spatial-temporal VAE decoder can produce numerous visually pleasing results. However, lacking guidance from the sketch, many results may not align with expected visual representations. Therefore, we introduce the sketch-aware encoder to guide this process. Specifically, the sketch-aware encoder adopts the same architecture as the VAE encoder, extracting multi-level texture features before each downsampling layer. Then, these features are added to the decoder at each corresponding level:
\begin{equation}
\mathcal{D}(z^\prime,\mathcal{E}^{\mathrm{s}}(s))_{i} = \mathcal{E}^{\mathrm{s}}(s)_i + \mathcal{D}(z^\prime)_i,
\end{equation}
where subscript $i$ denotes the feature level.
We train the sketch-aware encoder by minimizing a combination of an SSIM loss \cite{ssim} $\mathcal{L}_{\mathrm{sm}}$ for structure alignment, L1 loss $\mathcal{L}_\mathrm{1}$ for reconstruction fidelity, perceptual loss \cite{perceptual} $\mathcal{L}_{\mathrm{pc}}$ for sharp details, and KL loss \cite{klloss} $\mathcal{L}_{\mathrm{kl}}$ for regularization. The combined loss is calculated for each frame and then summed over all frames:
\begin{equation}
    \mathcal{L}_{\mathrm{vae}} = \sum_{i}^{N} \mathcal{L}_{\mathrm{sm}} + \lambda_1 \mathcal{L}_\mathrm{1} + \lambda_2 \mathcal{L}_{\mathrm{pc}} + \lambda_3 \mathcal{L}_{\mathrm{kl}},
\end{equation}
where $i$ indexes the frames. $\lambda_1=1$, $\lambda_2=0.1$, and $\lambda_3=1 \times 10^{-6}$ are hyper-parameters that are not sensitive to variations in a certain range.

\begin{figure*}[t]
   \centering
  \includegraphics[width=\linewidth]{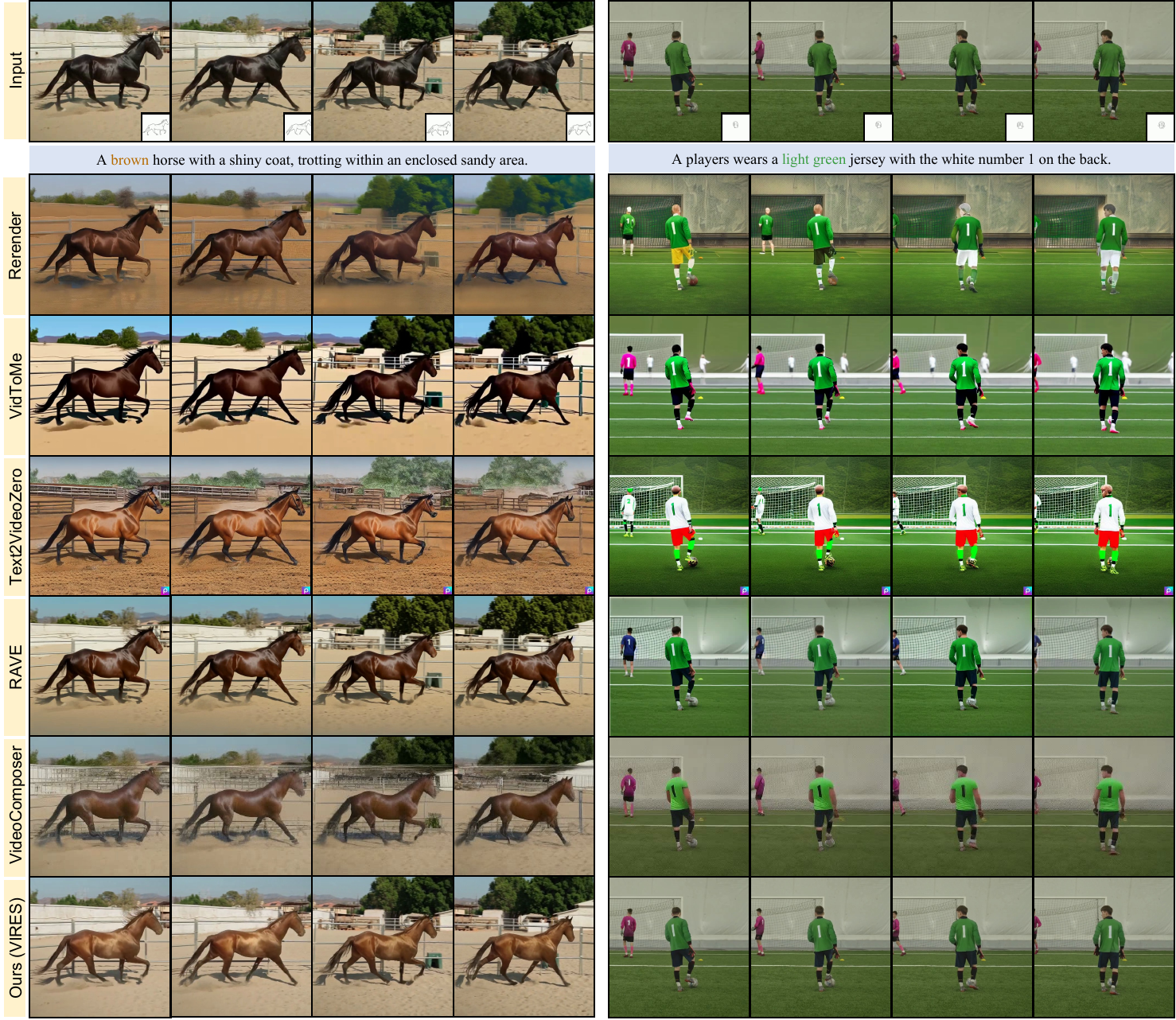}
  \caption{ Visual quality comparisons of video editing methods with text and sketch guidance.
  }
  \label{fig:comparison}
\end{figure*}

\section{Experiment}
\noindent \textbf{Training details.}
We initialize our spatial-temporal VAE and DiT backbone with pre-trained weights from OpenSora v1.2\footnote{https://huggingface.co/hpcai-tech/OpenSora-STDiT-v3} and train the VIRES model to repaint videos at a 512$\times$512 resolution. The training process consists of three steps: 
\textit{(i)} With the pre-trained spatial-temporal VAE encoder and decoder frozen, we train the sketch-aware encoder for 22K steps; \textit{(ii)} With both VAEs and the pre-trained DiT backbone frozen, we train the Sequential ControlNet, the standardized self-scaling, and output matrices of the self-attention layers for 35K steps; and \textit{(iii)} After augmenting the DiT backbone with the sketch attention, we train the model for an additional 45K steps.
All experiments are conducted on 8 H100 GPUs using the Adam optimizer~\cite{adam} with a learning rate of $1 \times 10^{-5}$.


\noindent \textbf{Evaluation Datasets.} We evaluate our approach on two datasets: our collected {\sc VireSet} and the widely-used DAVIS~\cite{davis}. 50 videos are randomly selected from each dataset for evaluation. The condition annotation process of the DAVIS is consistent with that of {\sc VireSet}, and details are presented in~\cref{sec:dataset}.

\subsection{Quantitative evaluation metrics}
We quantitatively evaluate performance using five metrics:
\textit{(i)} the \textbf{Peak Signal-to-Noise Ratio (PSNR)}~\cite{psnr} for the visual perceptual quality;
\textit{(ii)} 
the \textbf{Structural Similarity Index Measure (SSIM)}~\cite{ssim} for spatial structure consistency;
\textit{(iii)} 
the \textbf{Warp Error (WE)}, calculated by warping repainted video clips with the estimated optical flow, following VidToMe~\cite{vidtome} to evaluate motion accuracy;
\textit{(iv)} the \textbf{Frame Consistency (FC)}, measuring cosine similarity between consecutive frames for temporal consistency using the CLIP image encoder~\cite{clip};
and \textit{(v)} the \textbf{Text Consistency (TC)}, measuring cosine similarity between repainted video clips and text descriptions for text-video alignment using the VideoCLIP-XL~\cite{videoclipxl}.


\begin{table*}[t]
\caption{Quantitative experiment results of comparison and ablation. $\uparrow$ ($\downarrow$) means higher (lower) is better. All scores except PSNR are percentages. Throughout the paper, best performances are highlighted in \textbf{bold}.} 
\vspace{-2mm}
\begin{center}
{
    \setlength\tabcolsep{6pt}
    \centering
    \begin{adjustbox}{width={\textwidth},totalheight={\textheight},keepaspectratio}
    \begin{tabular}{l | c c c c c c c | c c c c c c c}  \toprule
    \multirow{2}{*}{Method} & \multicolumn{7}{c|}{\sc VireSet} & \multicolumn{7}{c}{DAVIS} \\
    & PSNR $\uparrow$ & SSIM $\uparrow$  & WE $\downarrow$  & FC $\uparrow$ & TC $\uparrow$ & VQE $\uparrow$ & TAE $\uparrow$  & PSNR $\uparrow$ & SSIM $\uparrow$  & WE $\downarrow$  & FC $\uparrow$ & TC $\uparrow$ & VQE $\uparrow$ & TAE $\uparrow$ \\ \midrule
    \multicolumn{15}{c}{\Large\bfseries Comparison with state-of-the-art methods} \cr \midrule
        Rerender & \text{11.85} & \text{35.21} & \text{5.77} & \text{90.40} & \text{15.78} & \text{6.40} & \text{5.60} & \text{11.79} & \text{29.67} & \text{8.01} & \text{89.75} & \text{15.56} & \text{7.20} & \text{8.00}
        \\
        
        VidToMe & \text{16.40} & \text{57.35} & \text{5.25} & \text{89.14} &\text{16.13} & \text{4.00} & \text{6.40} & \text{15.82} & \text{52.63} & \text{6.51} & \text{90.00} & \text{15.76} & \text{6.00} & \text{7.60}
        \\

        Text2VideoZero & \text{10.04} & \text{27.74} & \text{9.03} & \text{92.13} &\text{15.80} & \text{5.20} & \text{8.40} & \text{11.11} & \text{26.63} & \text{10.38} & \text{92.15} & \text{15.79} & \text{3.20} & \text{5.60}
        \\
        
        RAVE   & \text{18.99} & \text{70.81} & \text{6.18} & \text{92.17} & \text{15.89} & \text{18.00} & \text{23.60} & \text{19.08} & \text{68.35} & \text{8.74} & \text{91.99} & \text{15.81} & \text{23.20} & \text{25.60}
        \\

        VideoComposer & \text{16.57} & \text{53.99} & \text{5.15} & \text{89.82} & \text{15.49} & \text{12.00} & \text{12.80} & \text{17.15} & \text{49.70} & \text{6.52} & \text{90.24} & \text{15.70} & \text{9.60} & \text{10.80}
        \\ 
        Ours (VIRES) & \textbf{23.87} & \textbf{77.87} & \textbf{5.09} & \textbf{92.23} & \textbf{16.19} & \textbf{54.40} & \textbf{43.20} & \textbf{25.36} & \textbf{76.74} & \textbf{6.49} & \textbf{92.27} & \textbf{16.08} & \textbf{50.80} & \textbf{42.40}   \\ \midrule
    \multicolumn{15}{c}{\Large\bfseries Ablation study} \cr \midrule
        \textit{W/o} SSS & \text{22.72} & \text{75.81} & \text{5.17} & \text{91.94} & \text{16.13} & \text{N/A} & \text{N/A}& \text{24.59} & \text{75.66} & \text{6.55} & \text{91.88} & \text{16.02} & \text{N/A}
        & \text{N/A}
        \cr
        \textit{W/o} SS  & \text{23.09} & \text{76.31} & \text{5.13} & \text{91.99} & \text{16.16} & \text{N/A} & \text{N/A}& \text{24.95} & \text{75.84} & \text{6.52} & \text{91.96} & \text{16.04} & \text{N/A}
        & \text{N/A}
        \cr
        \textit{W/o} SA &  \text{22.89} & \text{76.45} & \text{5.12} & \text{92.01} & \text{16.08} & \text{N/A} & \text{N/A}& \text{24.86} & \text{76.06} & \text{6.54} & \text{91.92} & \text{15.99} & \text{N/A}
        & \text{N/A}
        \cr
        \textit{W/o} SE &  \text{22.94} & \text{76.13} & \text{5.17} & \text{91.80} & \text{16.11} & \text{N/A} & \text{N/A} & \text{24.72} & \text{75.78} & \text{6.58} & \text{91.89} & \text{16.02} & \text{N/A}
        & \text{N/A}
        \cr\bottomrule
    
    \end{tabular}\label{tab:comparison}
    \end{adjustbox}
}
\end{center}
\vspace{-2mm}
\end{table*}
\renewcommand{\arraystretch}{1} 

\subsection{Comparison with state-of-the-art methods}
We compare our approach with state-of-the-art video editing methods for generating 51-frame video clips, including VideoComposer~\cite{videocomposer}, Text2Video-Zero~\cite{text2videozero}, Rerender~\cite{renderavideo}, VidToMe~\cite{vidtome}, and RAVE~\cite{rave} using their publicly released models with default configurations. 

\noindent \textbf{Qualitative comparisons.}
We present visual quality comparison results with aforementioned methods \cite{videocomposer,text2videozero, renderavideo, vidtome, rave}. As shown in~\cref{fig:comparison}, the left column edits instance appearance using text descriptions while preserving structure with extracted sketches. Conversely, the right column maintains its appearance using annotated text descriptions and edits structure via sketch modifications.
As a result, Rerender \cite{renderavideo} produces excessively smooth textures (\eg, \cref{fig:comparison} left, the unnatural smoothness of the horse and tree).
VidToMe~\cite{vidtome} generates a cartoonish effect (\eg, \cref{fig:comparison} left, the stylized appearance of the entire frame).
Text2Video-Zero~\cite{text2videozero} presents inaccurate color tone (\eg, \cref{fig:comparison} right, the white jersey of the player).
RAVE~\cite{rave} suffers from inter-frame flickering (\eg, \cref{fig:comparison} right, the player's neck flashing with frame brightness fluctuations).
VideoComposer~\cite{videocomposer} struggles to follow the target structure (\eg, \cref{fig:comparison} right column, the missing sleeves of the player).
In contrast, our VIRES faithfully repaint the video instance with the sketch and text guidance.

\noindent \textbf{Quantitative comparisons.}
We show comprehensive quantitative comparison results in~\cref{tab:comparison}, demonstrating that VIRES outperforms relevant state-of-the-art methods across all five quantitative metrics on both datasets. 
Specifically, VIRES achieves the best results in visual perceptual quality (PSNR), spatial structure consistency (SSIM), frame motion accuracy (WE), consecutive frame consistency (FC), and text description consistency (TC).

\noindent \textbf{User study.}
In addition to qualitative and quantitative comparisons, we conduct two user studies to evaluate human preference for our results:
\textit{(i)} \textbf{Visual Quality Evaluation (VQE):}
Participants are shown the edited results produced by VIRES and relevant video editing methods. They are asked to select the most visually pleasing video clip.
\textit{(ii)} \textbf{Text Alignment Evaluation (TAE):} 
Given a corresponding text description, participants are instructed to select the video clip from the same set of edited results that best matches the description.
For each experiment, we randomly select 10 samples from each dataset, and recruit 25 volunteers from Amazon Mechanical Turk (AMT) to provide independent evaluations. As shown in \cref{tab:comparison}, our model achieves the highest preference scores in both experiments.

\subsection{Ablation study}
\begin{figure}[t]
  \centering
  \hfill
  \includegraphics[width=\linewidth]{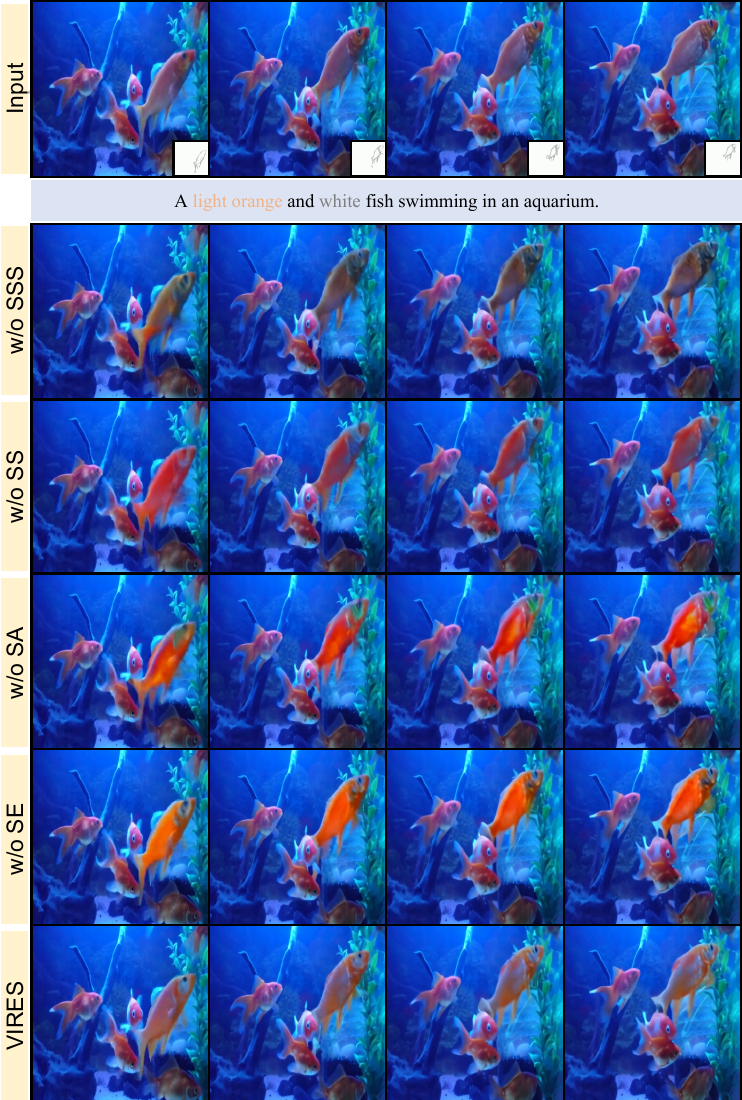}
  \caption{Ablation study results of different VIRES variants.}
  \label{fig:ablation}
  \vspace{-2mm}
\end{figure}

We discard several modules and establish four baselines to study the impact of the corresponding modules. The evaluation scores and repainted results of the ablation study are presented in \cref{tab:comparison} and \cref{fig:ablation}, respectively. 

\noindent \textbf{W/o Standardized Self-Scaling (SSS).} 
We replace the standardization self-scaling with simple addition when initially injecting the structure layout. This leads to distorted textures in the repainted results (\cref{fig:ablation} third column, an unusual texture appears around the fish's eye).

\noindent \textbf{W/o Self-Scaling (SS).}  
We discard the self-scaling, but preserve the standardization for the extracted structure layout before adding them to the DiT backbone. As a result, the model struggles to capture structure details (\cref{fig:ablation} second column, the fish head appears less detailed).

\noindent \textbf{W/o Sketch Attention (SA).} 
We remove the sketch attention in the DiT backbone, thereby removing the third stage of training. This prevents the model from accurately interpreting fine-grained sketch semantics (\cref{fig:ablation} fourth column, abrupt yellow patches appear on the fish's body).

\noindent \textbf{W/o Sketch-aware Encoder (SE).}
We disable the sketch-aware encoder that provides multi-level features to guide the decoding process, resulting in repainted results that are not well aligned with the expected visual representations (\cref{fig:ablation} first column, the fish body appearing blurry and flat).



\begin{figure}[t]
  \centering
  \hfill
  \includegraphics[width=\linewidth]{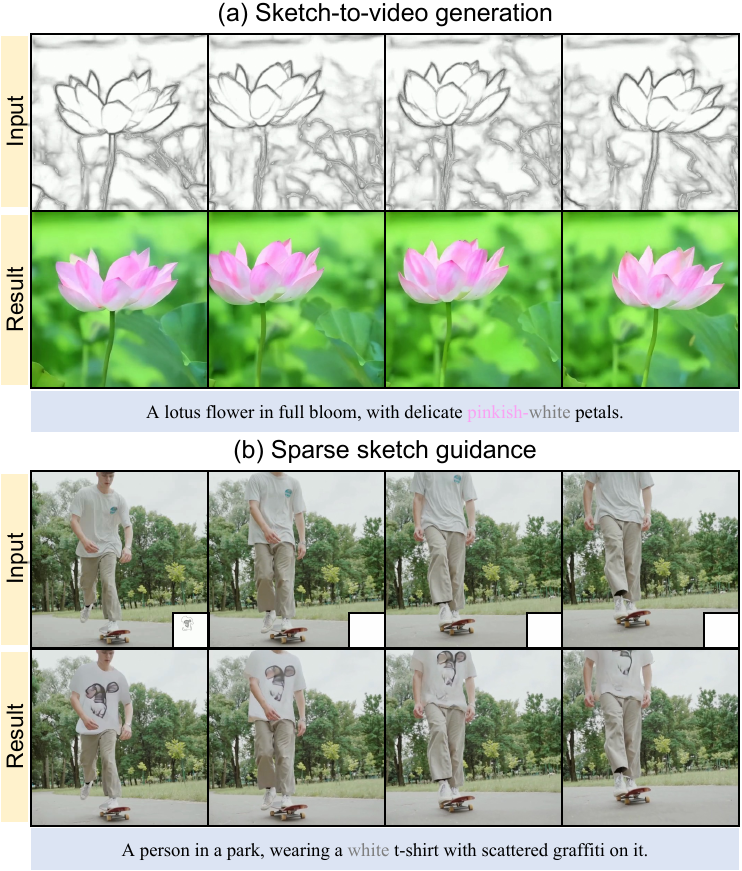}
  \caption{Examples of extended applications: (a) Generating an entire video clip with the sketch sequence. (b) Repainting the video instance with a single sketch frame.}
  \label{fig:discuss}
  \vspace{-2mm}
\end{figure}

\subsection{Application}
In addition to four application scenarios illustrated in~\cref{fig:teaser} (\ie, video instance repainting, video instance replacement, custom instance generation, and specified instance removal), we present additional three applications.

\noindent \textbf{Sketch-to-video generation.}
VIRES is capable of generating entire video clips from sketch sequences with text guidance, not limited to repainting specific instances. 
This is achieved by discarding the original video content and expanding the instance mask to the entire frame, enabling the training-free sketch-to-video generation (\eg, \cref{fig:discuss} first row, generating a realistic lotus flower from a provided sketch sequence).

\noindent \textbf{Sparse sketch guidance.}
To reduce user effort when repainting video instances, we incorporate sparse sketch guidance into the VIRES. Instead of requiring a full sketch sequence, we fine-tune the VIRES using two strategies: \textit{(i)} random dropout of sketch frames with 20\% probability; and splitting the video into $2^d$ intervals with 80\% probability, where $d \in \{0, \dots, 4\}$ is a randomly selected interval index. For each interval, only the first sketch frame is provided, with missing frames replaced by black images. As a result, VIRES can repaint video instances with even a single sketch frame (\eg, \cref{fig:discuss} second row, repainting the pattern on the person's clothes with the first sketch frame).


\noindent \textbf{Long-duration video repainting.}
Due to the computational resource limitation, processing the full long-duration video at once is challenging.
However, VIRES can effectively repaint long-duration videos by randomly unmasking the frames during training (\eg, the first $k$ frames).
Specifically, we initially repaint the first 51-frame video clip, leveraging the last 17 frames of this clip as a hint to iteratively repaint the subsequent 34-frame clips until the entire video is processed. We present the results in the Supp.

\section{Conclusion}
In this paper, we present the VIRES, a \textbf{V}ideo \textbf{I}nstance \textbf{RE}painting method with \textbf{S}ketch and text guidance. 
We build VIRES upon the pre-trained text-to-video generation model to maintain temporal consistency. We propose a Sequential ControlNet with a standardized self-scaling to effectively extract structure layouts and adaptively capture high-contrast sketch details.
We further augment the DiT backbone with the sketch attention to interpret and inject fine-grained sketch semantics, and introduce a sketch-aware encoder for structure alignment during decoding.  
For training and evaluating video editing methods, we contribute {\sc VireSet}. 
In addition to video instance repainting, replacement, generation, and removal, VIRES is applicable to sketch-to-video generation, sparse sketch guidance, and long-duration video repainting. 
Extensive experiments demonstrate that VIRES outperforms existing video editing methods and receives higher ratings from human evaluators.

\noindent \textbf{Limitation.}
Our VIRES is based on the DiT architecture~\cite{opensora}, which leads to higher computational requirements. For example, editing a 51-frame video at $512 \times 512$ resolution takes approximately 323 seconds on a single A6000 GPU, which is about 1.6 times slower than other diffusion-based methods~\cite{text2videozero, videocomposer}. We believe faster sampling methods will reduce the inference time in the future.

\section{Appendix}

\subsection{Variations of Sequential ControlNet } 
We present four typical architectures of Sequential ControlNet in~\cref{tab:variations}, where ``Conv'' denotes a convolutional layer, "Block" means a residual block, and "Down" is a downsampling layer. Numbers in brackets indicate the input and output channel dimensions, respectively. To determine the optimal architecture under constrained computational resources, we train these variations for 10K steps, excluding the sketch attention and sketch-aware encoder. Quantitative results on {\sc VireSet} are shown in \cref{tab:controlnet}, and the best-performing architecture is selected for our model.

\begin{table}[h]
\caption{Architecture of Sequential ControlNet variations.} 
\vspace{-2mm}
\begin{center}
{
    \setlength\tabcolsep{3pt}
    \centering
    \begin{adjustbox}{width={0.48\textwidth},totalheight={\textheight},keepaspectratio}
    \begin{tabular}{l | c c c c}  \toprule
    & 1 & 2 & 3 & 4 \\ \midrule
    \multirow{3}{*}{block\_1} & Conv (3, 72) & & Conv (3, 36) & Conv (3, 72) \\
    & Conv (72, 72) & & Conv (36, 36) & Conv (72,72) \\
    & Down (72, 144) & Down (3,64) & Down (36, 72) & Down (72, 144) \\ \midrule
    \multirow{5}{*}{block\_2} & Conv (144, 144) & Conv (64, 64) & & Conv (144,144) \\
    & Conv (144, 288) & Conv (64, 128) & & Conv (144, 288) \\
    & Conv (288, 288) & & & \\
    & Block (288, 288) & Block (128, 128) & Block (72, 72) & Block (288, 288) \\
    & Down (288, 576) & Down (128, 128) & Down (72, 288) & Down (288, 576) \\ \midrule
    \multirow{2}{*}{block\_3} & Block (576, 576) & Block (128, 256) & Block (288, 288) & Block (576, 576) \\
    & Down (576, 1152) & Down (256, 256) & Down (288, 1152) &  Down (576, 1152) \\ \midrule
    \multirow{4}{*}{block\_4} & Block (1152, 1152) & & & \\
    & Conv (1152, 1152) & Block (256, 256) & Conv (1152, 1152) & Conv (1152, 1152) \\
    & Conv (1152, 1152) &  & Conv (1152, 1152) & Conv (1152, 1152) \\
    & Conv (1152, 1152) & Conv (256, 1152) & Conv (1152, 1152) & Conv (1152, 1152) \\

    \bottomrule
    
    \end{tabular}\label{tab:variations}
    \end{adjustbox}
}
\end{center}
\vspace{-2mm}
\end{table}

\begin{table}[h]
\caption{Quantitative experiment results of Sequential ControlNet variations. All scores except PSNR are percentages.} 
\vspace{-2mm}
\begin{center}
{
    \setlength\tabcolsep{8pt}
    \centering
    \begin{adjustbox}{width={0.48\textwidth},totalheight={\textheight},keepaspectratio}
    \begin{tabular}{l | c c c c c}  \toprule
    \multirow{1}{*}{Method} & PSNR $\uparrow$ & SSIM $\uparrow$  & WE $\downarrow$  & FC $\uparrow$ & TC $\uparrow$ \\ \midrule
    Variation\_1 & \text{18.54} & \text{69.12} & \text{5.36} & \text{90.79} & \text{15.85}  \\
        
    Variation\_2 & \text{18.49} & \text{68.63} & \text{5.47} & \text{89.14} &\text{15.83} \\

    Vairation\_3 & \text{18.64} & \text{69.89} & \text{5.40} & \text{90.94} &\text{15.85}  \\ \midrule
        
    Ours (VIRES) & \textbf{19.93} & \textbf{72.85} & \textbf{5.33} & \textbf{91.07} & \textbf{15.87} \\ \bottomrule
    
    \end{tabular}\label{tab:controlnet}
    \end{adjustbox}
}
\end{center}
\vspace{-2mm}
\end{table}

\subsection{Validation of additional conditions} 
\begin{figure}[t]
   \centering
  \includegraphics[width=\linewidth]{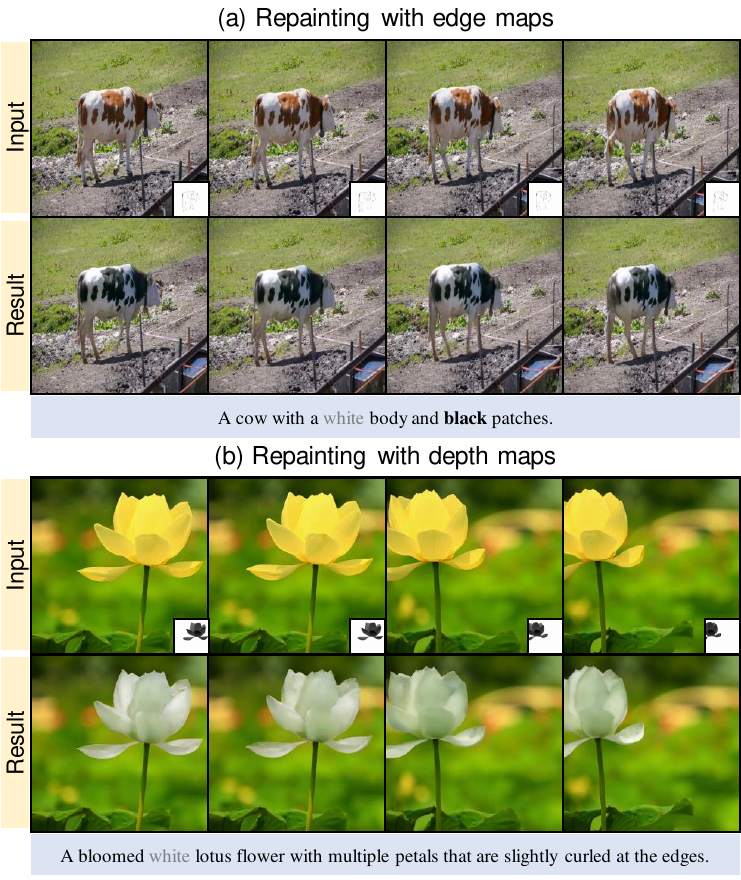}
  \caption{Examples of repainting with additional conditions.}
  \label{fig:edge_and_depth}
\end{figure}

VIRES is designed specifically for sketch-based video instance repainting, adopting the Standardized Self-Scaling (SSS) to extract condition features. To explore its generalization to other conditioning signals, we augment our dataset with edge maps~\cite{laplace} and depth maps~\cite{depthanything2} to provide the structure guidance. We then retrain VIRES from scratch on these conditions.  Due to limited computational resources, we train for 10K steps with edge maps as a preliminary investigation and 30K steps with depth maps due to its slower convergence. Finally, we compare two VIRES variants: one using SSS and the other using simple addition for feature extraction.
As shown in~\cref{tab:condition}, SSS provides considerable advantages for edge maps, but only marginal improvements for depth maps. We suggest that this difference arises because both sketch and edge maps have high-contrast transitions between black lines and the white background, allowing self-scaling to effectively capture structure details. In contrast, depth maps are smoother and lack such sharp transitions, limiting the benefits of the self-scaling operation. We further present the qualitative results in~\cref{fig:edge_and_depth}.

\begin{table}[h]
\caption{Quantitative experiment results of VIRES variants using edge and depth maps. All scores except PSNR are percentages.} 
\vspace{-2mm}
\begin{center}
{
    \setlength\tabcolsep{8pt}
    \centering
    \begin{adjustbox}{width={0.48\textwidth},totalheight={\textheight},keepaspectratio}
    \begin{tabular}{l | c c c c c}  \toprule
    \multirow{1}{*}{Method} & PSNR $\uparrow$ & SSIM $\uparrow$  & WE $\downarrow$  & FC $\uparrow$ & TC $\uparrow$ \\ \midrule
    \multicolumn{6}{c}{Repainting with edge maps} \\

    W/o SSS & \text{19.43} & \text{72.01} & \text{6.31} & \text{90.04} &\text{15.83} \\ 
    W/ SSS & \textbf{20.48} & \textbf{74.28} & \textbf{6.00} & \textbf{90.49} & \textbf{16.16}  \\ \midrule

    \multicolumn{6}{c}{Repainting with depth maps} \\
    W/o SSS & \text{18.04} & \text{66.16} & \text{5.86} & \text{91.28} & \text{15.98} \\ 
    W/ SSS & \textbf{18.39} & \textbf{67.28} & \textbf{5.80} & \textbf{91.32} & \textbf{16.14}  \\ \bottomrule
    
    \end{tabular}\label{tab:condition}
    \end{adjustbox}
}
\end{center}
\end{table}

\subsection{Robustness of sparse conditions}
VIRES allows users to provide sparse sketch guidance for video instance repainting, minimizing user effort. 
To investigate the impact of sketch guidance sparsity on repainting performance, we evaluate VIRES on the {\sc VireSet}, using varying interval indices $d \in \{0,\dots,4\}$, corresponding to $D=2^d$ sketch frames, and a full set of $D=51$. As shown in~\cref{tab:sparse}, even with a single sketch frame, VIRES can produce high-quality results (PSNR and SSIM) with robust temporal consistency (WE and FC).

\begin{table}[t]
\caption{Quantitative experiment results of VIRES using sketch guidance of varying sparsity. All scores except PSNR are percentages.} 
\vspace{-2mm}
\begin{center}
{
    \setlength\tabcolsep{8pt}
    \centering
    \begin{adjustbox}{width={0.48\textwidth},totalheight={\textheight},keepaspectratio}
    \begin{tabular}{l | c c c c c}  \toprule
    \multirow{1}{*}{Method} & PSNR $\uparrow$ & SSIM $\uparrow$  & WE $\downarrow$  & FC $\uparrow$ & TC $\uparrow$ \\ \midrule
    $D=1$ & \text{21.87} & \text{73.41} & \text{5.12} & \text{92.21} &\text{16.15} \\ 
    $D=2$ & \text{22.64} & \text{74.69} & \text{5.14} & \text{92.21} & \text{16.17} \\
    $D=4$ & \text{23.24} & \text{76.00} & \text{5.13} & \text{92.18} & \text{16.16} \\ 
    $D=8$ & \text{23.36} & \text{76.51} & \text{5.14} & \text{92.19} & \text{16.18}  \\
    $D=16$ & \text{23.67} & \text{77.15} & \text{5.12} & \text{92.19} & \text{16.19}\\
    $D=51$ & \textbf{23.87} & \textbf{77.87} & \textbf{5.09} & \textbf{92.23} & \textbf{16.19}
    \\ \bottomrule
    \end{tabular}\label{tab:sparse}
    \end{adjustbox}
}
\end{center}
\end{table}

\subsection{Compatibility with DiT backbone}
In this paper, we build VIRES upon the pre-trained OpenSora v1.2~\cite{opensora}. Given that many text-to-video models~\cite{allegro, cogvideox} share a similar DiT backbone architecture, our proposed modules offer potential compatibility with these approaches, including the Sequential ControlNet for layout initialization (Sec. \textcolor{red}{4.2}), the standardized self-scaling for details capture (Sec. \textcolor{red}{4.2}), the sketch attention (Sec. \textcolor{red}{4.3}) for semantic injection, and the sketch-aware encoder for structure alignment (Sec. \textcolor{red}{4.4}). 
We believe our work will inspire further research on guiding pre-trained text-to-video models and open new avenues for conditional video repainting.

\subsection{Organization of supplementary video}
\begin{figure*}[t]
   \centering
  \includegraphics[width=\linewidth]{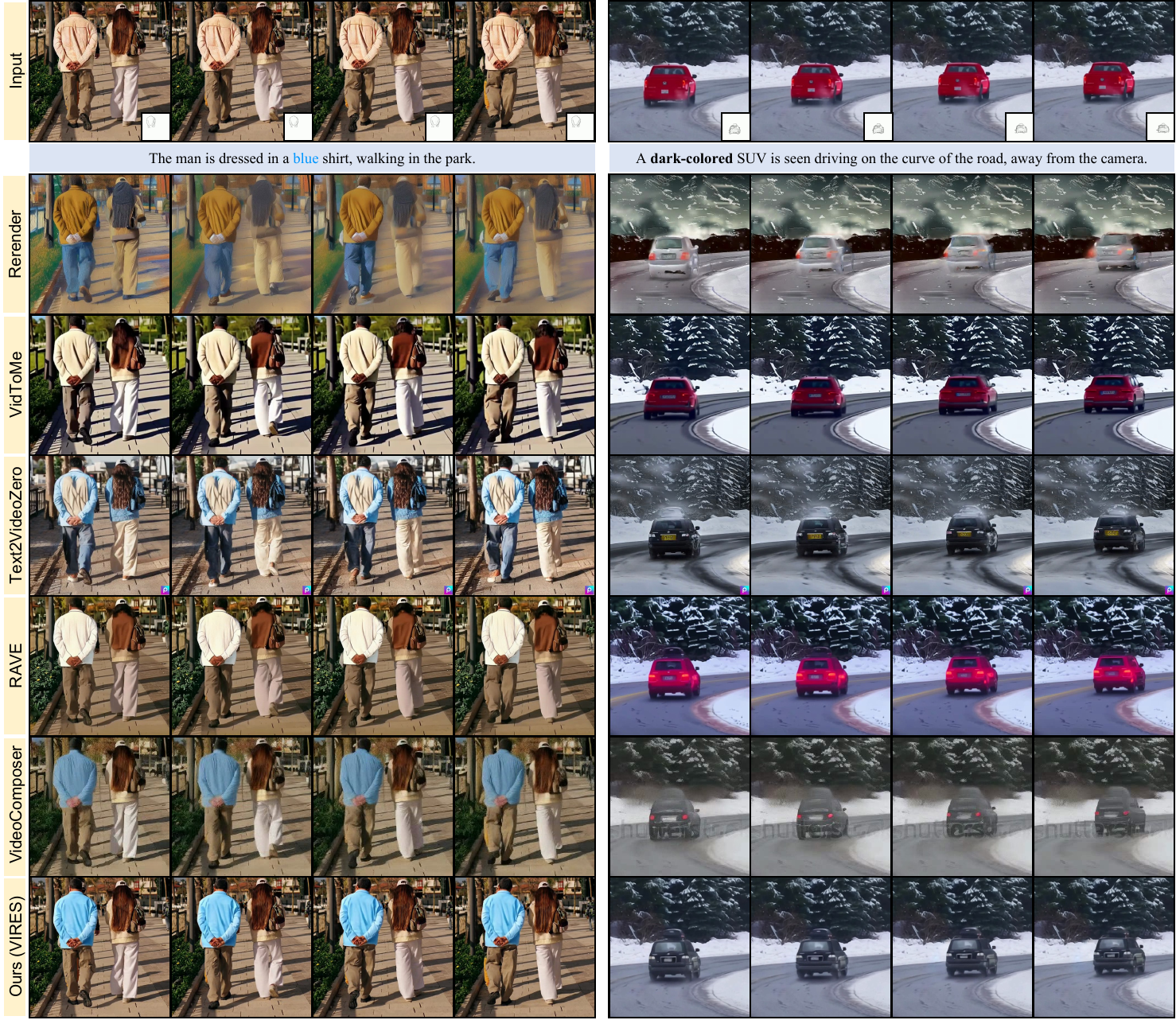}
  \caption{Typical application scenarios. \textbf{Left:} Video instance repainting. \textbf{Right:} Video instance replacement.}
  \label{fig:teaser1}
\end{figure*}

\begin{figure*}[t]
   \centering
  \includegraphics[width=\linewidth]{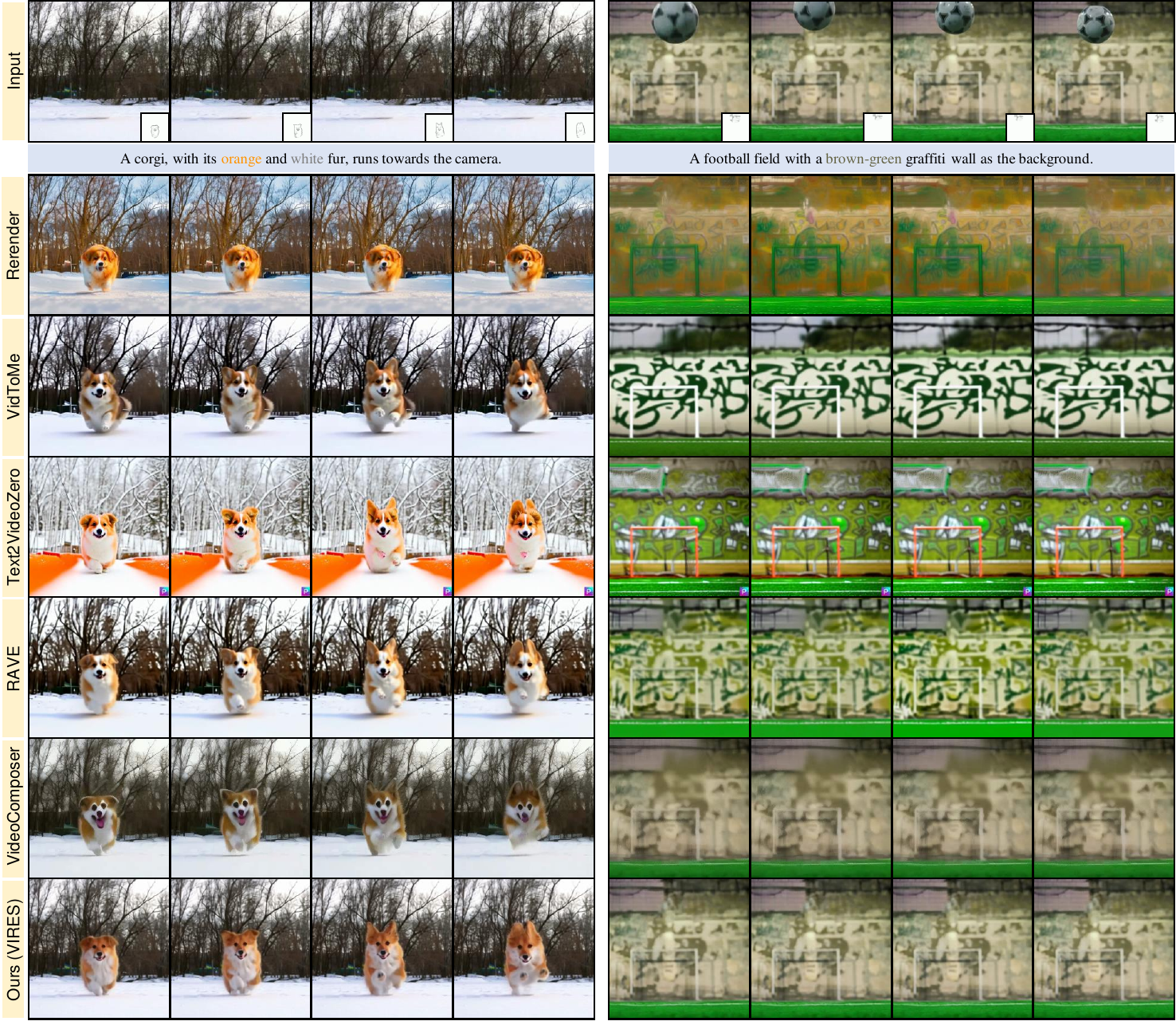}
  \caption{Typical application scenarios. \textbf{Left:} Custom instance generation. \textbf{Right:} Specified instance removal.}
  \label{fig:teaser2}
\end{figure*}

\begin{figure*}[t]
   \centering
  \includegraphics[width=\linewidth]{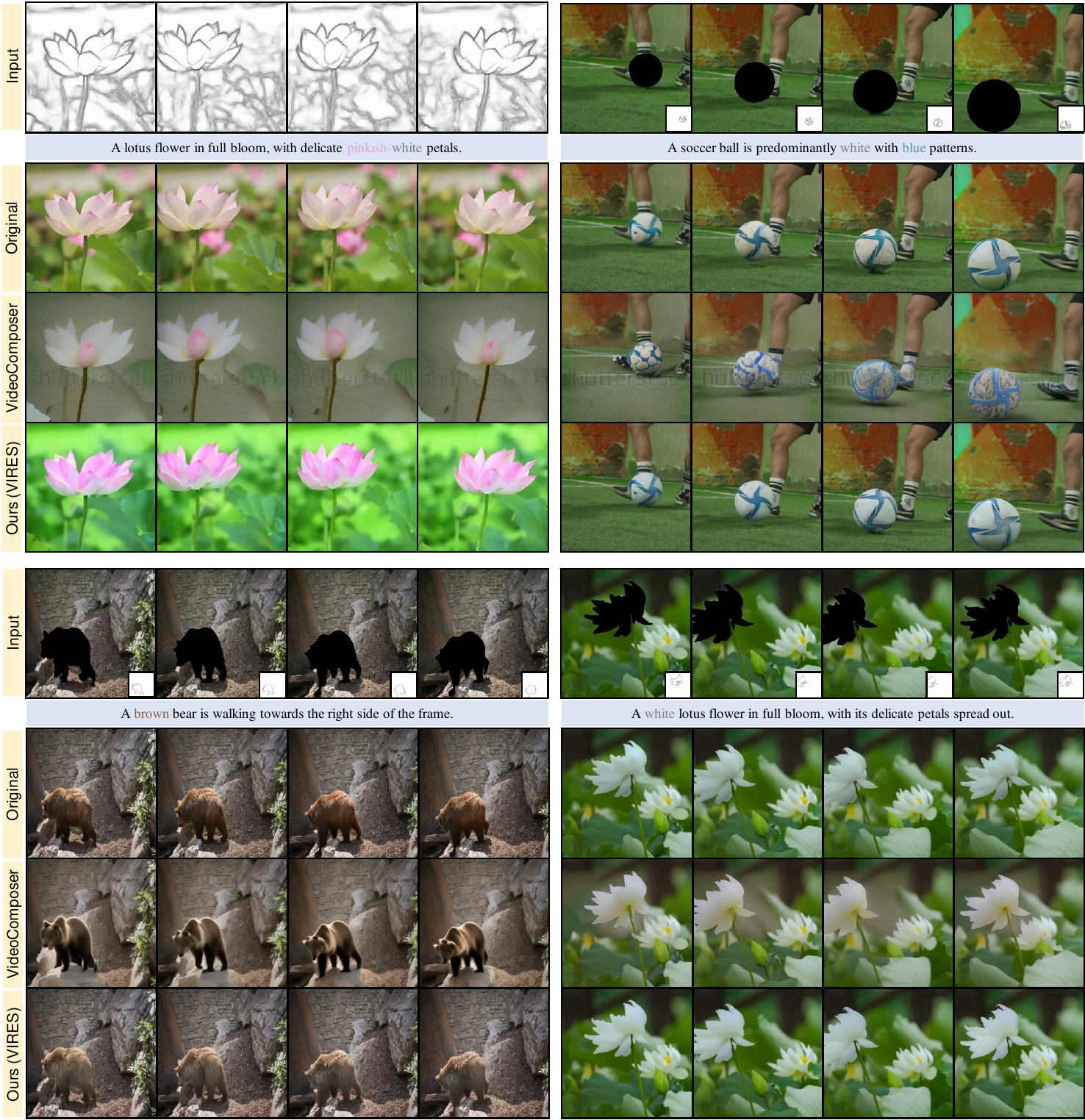}
  \caption{Sketch-to-video generation and inpainting. \textbf{Topleft:} Sketch-to-video generation. \textbf{Others:} Sketch-to-video inpainting.}
  \label{fig:inpainting}
\end{figure*}

\begin{figure*}[t]
   \centering
  \includegraphics[width=\linewidth]{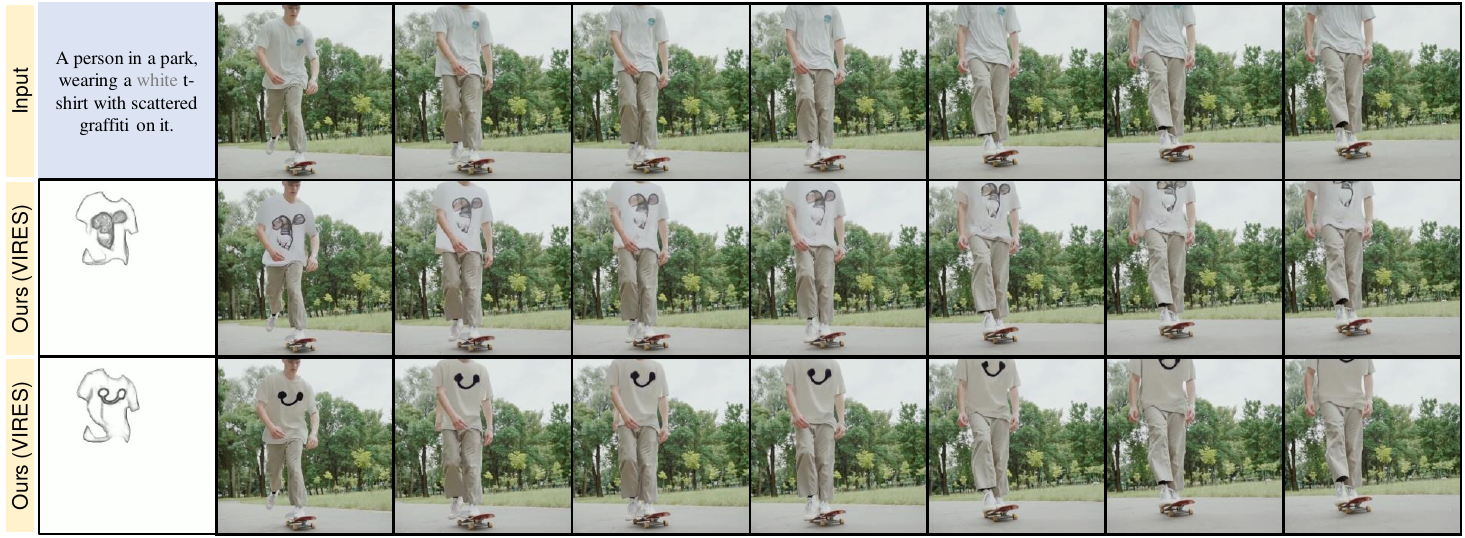}
  \caption{Sparse sketch guidance. Repainting the video using different first sketch frames. \textbf{Left:} First variation. \textbf{Right}: Second variation.}
  \label{fig:sparse}
\end{figure*}

\begin{figure*}[t]
   \centering
  \includegraphics[width=\linewidth]{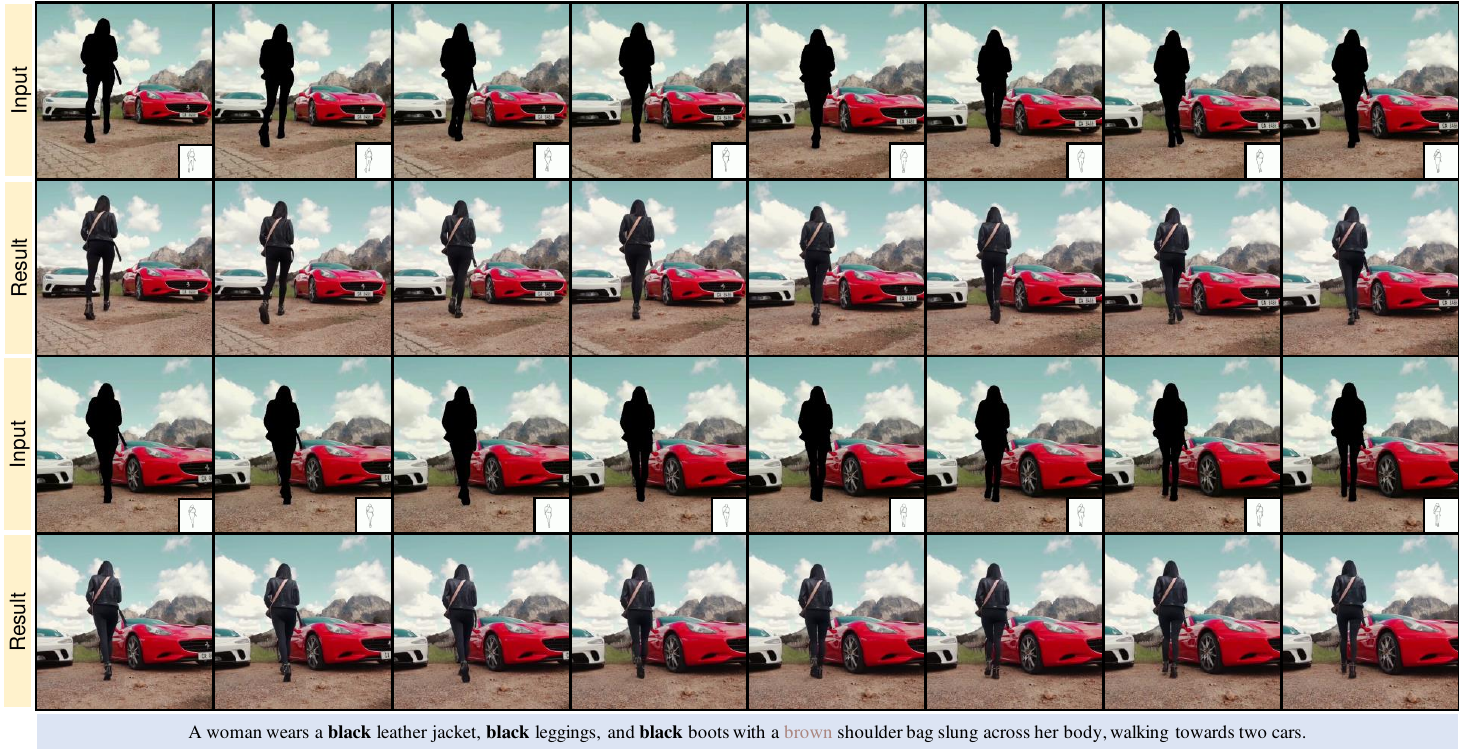}
  \caption{Long-duration video repainting. Restoring the damaged video to clearly depict a realistic female character.}
  \label{fig:longvideo}
\end{figure*}

We provide a supplementary video to dynamically showcase our repainting results. The video is structured as follows:
\textit{(i)} \textbf{Typical application scenarios:} We demonstrate four typical repainting scenarios and compare our results with relevant methods~\cite{renderavideo, vidtome, text2videozero, rave, videocomposer}. Instance repainting/replacement results are shown in~\cref{fig:teaser1}, and instance generation/removal results are in~\cref{fig:teaser2}. 
\textit{(ii)} \textbf{Sketch-to-video generation and inpainting:} We demonstrate sketch-to-video generation and conditional video inpainting, comparing our results with VideoComposer~\cite{videocomposer}, as it is the only relevant method that supports this functionality. Results are shown in~\cref{fig:inpainting}.
\textit{(iii)} \textbf{Sparse sketch guidance}: We showcase sparse sketch guidance, repainting two distinct variations of the same video using only two different first sketch frames. This functionality is not supported by existing methods. Results are shown in~\cref{fig:sparse}.
\textit{(iv)} \textbf{Long-duration video repainting:} We demonstrate repainting on a long-duration (13-second) video, with representative frames presented in~\cref{fig:longvideo}.
\textit{(v)} \textbf{Comparison and ablation study:} Finally, the video includes comparisons with other methods (Sec.~\textcolor{red}{5.1}) and additional ablation studies (Sec.~\textcolor{red}{5.2}).
To improve visual clarity and facilitate detailed comparison, the video playback speed is halved ($2\times$ slower).

\clearpage
\clearpage
{
    \small
    \setlength{\itemsep}{0pt} 
    \bibliographystyle{ieeenat_fullname}
    \bibliography{main}
}
\end{document}

%% file: preamble.tex
\usepackage[dvipsnames]{xcolor}
